\makeatletter\def\graphicscache@inhibit{true}\makeatother

\documentclass[a4paper, 10pt, conference]{ieeeconf}      %

\IEEEoverridecommandlockouts                              %

\usepackage[utf8]{inputenc}
\usepackage[T1]{fontenc}
\usepackage{textcomp}
\usepackage{graphicx} %
\usepackage{graphicscache}

\usepackage{pgfplots}
\pgfplotsset{compat=newest}
\usepackage{subfig}
\usepackage{multirow}
\usepackage{tabularx}
\usepackage{threeparttable}
\usepackage{booktabs}
\usepackage[hidelinks]{hyperref}
\usepackage{array}
\usepackage{float}
\usepackage{graphbox}
\usepackage{tikz,tikz-3dplot}
\usetikzlibrary{arrows,arrows.meta,automata,backgrounds,calc,chains,%
decorations.markings,decorations.pathreplacing,decorations.pathmorphing,%
matrix,positioning,shapes,shapes.geometric,shapes.symbols,spy,trees,tikzmark}
\usepackage{balance}
\usepackage[binary-units=true,product-units=single,per-mode=symbol,range-units=single,detect-all]{siunitx}
\DeclareSIUnit\pixel{px}
\usepackage{amsmath} %
\usepackage{amssymb}  %
\usepackage{bm}
\usepackage{acronym}
\usepackage{tablefootnote}

\definecolor{bg_color}{RGB}{95,95,95}

\let\vec\bm

\DeclareMathOperator*{\argmax}{arg\,max}

\newcommand{\reffig}[1]{Fig.~\ref{#1}}
\newcommand{\reftab}[1]{Tab.~\ref{#1}}
\newcommand{\refsec}[1]{Sec.~\ref{#1}}
\newcommand{\refeq}[1]{Eq.~\ref{#1}}
\newcommand{\etal}{et al.~}
\newcommand{\citep}[1]{(\cite{#1})}

\newcommand{\wrt}{~w.r.t.~}
\newcommand{\ie}{i.e.,\ }
\newcommand{\eg}{e.g.,\ }
\newcommand{\cf}{cf.\ }

\usepackage{adjustbox}
\newcolumntype{R}[2]{%
    >{\adjustbox{angle=#1,lap=\width-(#2)}\bgroup}%
    l%
    <{\egroup}%
}
\newcommand*\rot{\multicolumn{1}{R{40}{1em}}}%
\newcolumntype{L}[1]{>{\raggedright\let\newline\\\arraybackslash\hspace{0pt}}m{#1}}

\title{\LARGE \bf
Real-Time Multi-Modal Semantic Fusion on Unmanned Aerial Vehicles
}

\author{Simon Bultmann$^{*}$, Jan Quenzel$^{*}$, and Sven Behnke%
\thanks{\hspace{-2.2ex}$^{*}$: equal contribution.}%
\thanks{This work has been supported by the German Federal Ministry of Education and Research (BMBF) in the project ``Kompetenzzentrum: Aufbau des Deutschen Rettungsrobotik-Zentrums (A-DRZ)''}%
\thanks{Institute for Computer Science VI, Autonomous Intelligent Systems, University of Bonn, Friedrich-Hirzebruch-Allee 8, 53115 Bonn, Germany,
		{\tt\small \{bultmann,quenzel,behnke\}@ais.uni-bonn.de}%
}
\thanks{978-1-6654-1213-1/21/\$31.00 \textcopyright 2021 IEEE}
}

\begin{document}

\maketitle

\begin{tikzpicture}[remember picture,overlay]
\node[anchor=north west,align=left,font=\sffamily,yshift=-0.2cm] at (current page.north west) {%
  Accepted for: 10th European Conference on Mobile Robots (ECMR), Bonn, Germany, September 2021. 
};
\end{tikzpicture}%
\vspace{-1em}

\thispagestyle{empty}
\pagestyle{empty}

\begin{abstract}
Unmanned aerial vehicles (UAVs) equipped with multiple complementary sensors have tremendous potential for fast autonomous or remote-controlled semantic scene analysis, \eg for disaster examination.

In this work, we propose a UAV system for real-time semantic inference and fusion of multiple sensor modalities.
Semantic segmentation of LiDAR scans and RGB images, as well as object detection on RGB and thermal images, run online onboard the UAV computer using lightweight CNN architectures and embedded inference accelerators. We follow a late fusion approach where semantic information from multiple modalities augments 3D point clouds and image segmentation masks while also generating an allocentric semantic map.

Our system provides augmented semantic images and point clouds with $\approx\,$\SI{9}{\hertz}. We evaluate the integrated system in real-world experiments in an urban environment.
\end{abstract}

\section{Introduction}
\label{sec:Introduction}
Semantic scene understanding is an important prerequisite for solving many tasks with unmanned aerial vehicles (UAVs) or other mobile robots, \eg for disaster examination in search and rescue scenarios.
Modern robotic systems employ a multitude of different sensors to perceive their environment, \eg 3D LiDAR, RGB(-D) cameras, and thermal cameras, that capture complementary information about the environment. A LiDAR provides accurate range measurements independent of the lighting conditions, while cameras provide dense texture and color in the visible spectrum. Thermal cameras are especially useful in search and rescue missions as they detect persons or other heat sources regardless of lighting or visibility conditions. 
The combination of all these sensor modalities enables a complete and detailed interpretation of the environment. A semantic map aids inspection tasks~\cite{nguyen_mavnet_2019}, perception-aware path planning~\cite{bartolomei_perception-aware_2020}, and increases robustness and accuracy of simultaneous localization and mapping (SLAM) through the exclusion of dynamic objects during scan matching~\cite{chen_suma_2019}.

In this work, we propose a framework for online multi-modal semantic fusion onboard a UAV combining 3D LiDAR range data with 2D RGB and thermal images.
An embedded inference accelerator and the integrated GPU (iGPU) run inference online, onboard the UAV for mobile, optimized CNN architectures to obtain pixel- resp. pointwise semantic segmentation for RGB images and LiDAR scans, as well as object bounding box detections on RGB and thermal images.
We aggregate extracted semantics for two different output views: A fused segmentation mask for the RGB image which can, \eg be streamed to the operator for direct support of their situation awareness, and a semantically labeled point cloud, providing a 3D semantic scene view which is further integrated into an allocentric map. This late fusion approach is beneficial for multi-rate systems, increasing adaptability to changing sensor configurations and enabling pipelining for efficient hardware usage.

In summary, our main contributions are:
\begin{itemize}
\item the adaptation of efficient CNN architectures for image and point cloud semantic segmentation and object detection for processing onboard a UAV using embedded inference accelerator and iGPU,
\item the fusion of point cloud, RGB, and thermal modalities into a joint image segmentation mask and a semantically labeled 3D point cloud, 
\item temporal multi-view aggregation of the semantic point cloud and integration into an allocentric map, and
\item evaluation of the proposed integrated system with real-world UAV experiments.
\end{itemize}

\begin{figure}[t]
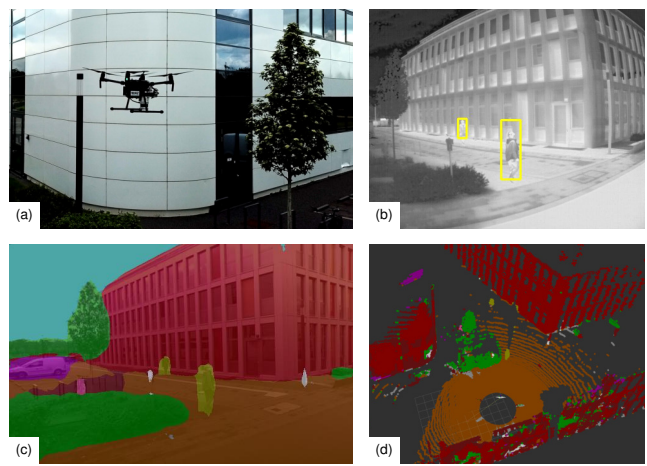

	\centering
		\begin{tikzpicture}
			\node[inner sep=0,anchor=south west] (image1) at (0, 0) {\includegraphics[height=2.9cm,trim=0 10 0 0, clip]{figures/copter_1.png}};
			\node[inner sep=0,anchor=south west,xshift=0.2cm] (image2) at (image1.south east) {\includegraphics[height=2.9cm]{figures/scenes/scene0_thermal_dets.jpg}};
			\node[inner sep=0,anchor=north west,yshift=-0.2cm] (image3) at (image1.south west) {\includegraphics[height=2.9cm,trim= 0 0 48 0,clip]{figures/scenes/scene0_fused_segm.jpg}};
			\node[inner sep=0,anchor=north west,yshift=-0.2cm] (image4) at (image2.south west) {\includegraphics[height=2.9cm,trim= 250 0 122 0, clip]{figures/scenes/scene0_pointcloud_1620389975651920436.jpg}};
			
			\node[label,scale=0.75, anchor=south west, rectangle, fill=white, align=center, font=\scriptsize\sffamily] (n_0) at (image1.south west) {(a)};
			\node[label,scale=0.75, anchor=south west, rectangle, fill=white, align=center, font=\scriptsize\sffamily] (n_1) at (image2.south west) {(b)};
			\node[label,scale=0.75, anchor=south west, rectangle, fill=white, align=center, font=\scriptsize\sffamily] (n_2) at (image3.south west) {(c)};
			\node[label,scale=0.75, anchor=south west, rectangle, fill=white, align=center, font=\scriptsize\sffamily] (n_3) at (image4.south west) {(d)};
		\end{tikzpicture}
	\caption{Semantic perception with UAV (a): (b) Person detections in thermal camera, (c) fused image segmentation and (d) point cloud segmentation.}
	\vspace{-1.5em}
	\label{fig:teaser}
\end{figure}
 
\section{Related Work}
\label{sec:Related_Work}
\paragraph*{Mobile Lightweight Vision CNNs}
Lightweight CNN architectures for computer vision tasks that are efficient and perform well on systems with restricted computational resources, \eg on mobile or embedded platforms, have become of increasing research interest in recent years. The MobileNet architectures~\cite{mobilenetv2_2018,mobilenetv32019} replace classical backbone networks such as ResNets~\cite{he_deep_2016} in many vision models while decreasing the number of parameters and the computational cost significantly, \eg by replacing standard convolutions with depthwise-separable convolutions---at the expense of a slight reduction in accuracy.

In object detection, single-stage architectures such as SSD~\cite{liu_ssd_2016} or YOLO~\cite{Redmon_YOLO_2016} have proven to be efficient in mobile applications through the use of predefined anchors instead of additional region proposal networks. Zhang \etal\cite{Zhang_2019_ICCV} further optimize YOLOv3 for usage onboard a UAV. However, the authors evaluate their network, called Slim-YOLOv3, only on a powerful discrete GPU which is not feasible for integration onboard a typical UAV.

Recently, Xiong \etal introduced MobileDets~\cite{xiong_mobiledets_2021} based on the SSD architecture with MobileNet~v3 backbone and optimized for embedded inference accelerators such as the Google EdgeTPU~\cite{edgetpu_usb}, which we employ onboard our UAV.

For semantic image segmentation, efficient architectures for inference onboard UAVs have mostly been proposed for specific applications, such as UAV tracking and visual inspection~\cite{nguyen_mavnet_2019} or weed detection for autonomous farming~\cite{weedNet2018}. %
The DeepLab~v3+ architecture~\cite{deeplabv3plus2018} shows state-of-the-art performance on large, general datasets while including elements of the MobileNet architectures such as depthwise-separable convolutions for efficient computation. In our work, we employ a DeepLab~v3+ model with MobileNet~v3 backbone for image segmentation.

For point cloud semantic segmentation, projection-based methods~\cite{cortinhal_salsanext_2020,milioto_rangenet_2019,xu_squeezessegv3_2020} utilize the image-like 2D structure of rotating LiDARs. This allows performing efficient 2D-convolutions and using well-known techniques from image segmentation. The downside of this approach is the restriction to single LiDAR scans in contrast to larger aggregated point clouds~\cite{qi2021offboard}. In this work, we adopt the SalsaNext architecture~\cite{cortinhal_salsanext_2020}, trained on the large-scale SemanticKITTI dataset~\cite{behley2019iccv} for autonomous driving, as it shows a good speed-accuracy trade-off.

\paragraph*{Semantic Mapping}
Many high-level robotic tasks benefit from or require semantic information about the environment.
For this, semantic mapping systems build an allocentric semantic environment model, anchored in a fixed, global coordinate frame.

Maturana \etal\cite{maturana_looking_2017} propose to extend existing digital elevation maps (DEM) with the detection of cars from UAVs.

SemanticFusion~\cite{mccormac_semanticfusion_2017} models surfaces as surfels where a Gaussian approximates the point distribution. For SLAM, this approach builds on ElasticFusion~\cite{whelan2015elasticfusion} and requires an RGB-D camera. A CNN generates pixel-wise class probabilities from the color image. Their fusion takes a Bayesian approach assuming that individual segmentations are independent and stores all class probabilities per surfel. 

The LiDAR surfel mapping SuMa++ by Chen \etal\cite{chen_suma_2019} uses a surfel's semantic class to further improve the registration accuracy by penalizing inter-class associations during scan matching and surfel update. Here, the projection-based RangeNet++~\cite{milioto_rangenet_2019} provides per point class probabilities.

With Recurrent-OctoMap, Sun \etal\cite{sun_recurrent-octomap_2018} aim at long-term mapping within changing environments. Here, each cell within the OctoMap~\cite{hornung2013octomap} contains an LSTM fusing point-wise semantic features and all LSTMs share weights.

Rosu \etal\cite{rosu2020semi} extract a mesh from an aggregated point cloud. Projection of mesh faces into images enables the transfer from image segmentation to a semantic texture. While projection and fusion happen in real-time, the required mesh generation and UV-unwrapping are done in pre-processing. Since only the argmax class is of interest and to meet GPU memory limitations, the sparse texture retains a small number of classes with high probability and discards all others. 

\paragraph*{Multi-Modal Semantic Fusion}
Mobile robotic systems, such as UAVs or self-driving cars, are often equipped with both camera and LiDAR sensors, as they provide complementary information. A LiDAR accurately measures ranges sparsely and independent of lighting conditions while cameras provide dense textures and colors. Hence, research focused on the fusion of camera and LiDAR for 3D detection and segmentation in the context of autonomous driving.

Xu \etal propose PointFusion~\cite{Xu_pointfusion_2018}, a two-stage pipeline for 3D bounding-box detection. It first processes a LiDAR scan with PointNet~\cite{Qi_2017_pointnet} and an image with ResNet independently, before fusing them on feature level with an MLP.

Meyer \etal\cite{Meyer_2019_CVPR_Workshops} take a similar sequential feature-level fusion approach, addressing both 3D object detection and dense segmentation. The feature-level fusion requires representing the LiDAR scan as a range image. Range and color image are cropped to the overlapping field-of-view (FoV), reducing the \SI{360}{\degree} horizontal FoV of the LiDAR to only \SI{90}{\degree}.

Vora \etal\cite{Vora_2020_CVPR} propose to in-paint point clouds with image segmentation by projecting LiDAR points into the image and assigning segmentation scores of the pixels. A 3D object detection network then processes the augmented point cloud.

In our work, different networks process LiDAR scan, RGB, and thermal images individually. We adopt a pro\-jec\-tion-based approach similar to~\cite{Vora_2020_CVPR} for multi-modal fusion in a multi-rate system. When multiple modalities are available, we use a linear combination to merge class probabilities from different sensors. Our mapping integrates augmented point clouds in a sparse voxel hash-map with per voxel full class probabilities. We adapt the Bayesian fusion of SemanticFusion to the logarithmic form for higher precision and greater numerical stability.
While being less popular in recent work, such a late fusion approach has important practical advantages for deployment on an integrated robotic system. Different FoVs and data rates are easy to handle and intermediate results, such as image segmentation or detections, are useful as stand-alone outputs.
Pipelining also allows reducing latency of sequentially executed individual networks during online operation. Furthermore, the smaller, simpler standard architectures of individual networks are easier to adapt and optimize for the embedded inference accelerators employed in this work.

\section{Our Method}
\label{sec:method}
\subsection{System Setup}
\begin{figure}[t]
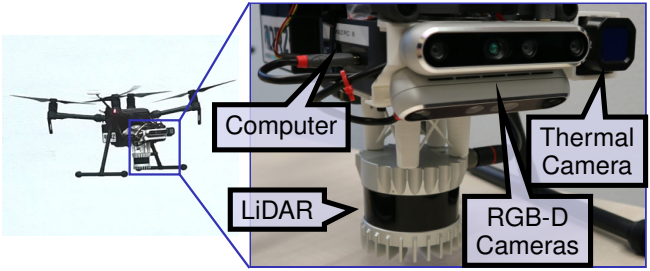

  \centering
  \resizebox{1.0\linewidth}{!}{
  \begin{tikzpicture}
  	[boxstyle/.style={font=\footnotesize\sffamily,black,fill=blue!20!white,fill opacity=0.8,text opacity=1,text=black,draw,ultra thick,align=center,rectangle callout}]
  		\definecolor{red}{rgb}{0.7,0.0,0.0}
    	\definecolor{blue}{rgb}{0.2,0.2,0.7}
		\node[anchor=north west, inner sep=0] (mav) at (0, 0) {\includegraphics[trim=45px 0px 25px 0px,clip,width=0.3\linewidth]{figures/copter_white.png}};
  		\node[draw,thick,line width=0.5mm,blue,anchor=center,xshift=3.6cm, inner sep=0] (closeup) at (mav.center) {\includegraphics[trim=0px 0px 0px 0px,clip,width=0.5\linewidth]{figures/sensors.png}};
  		
  		\draw[line width=0.25mm, blue] (1.4, -1.55) rectangle ++(0.55,0.6);
		\draw[line width=0.25mm, blue] (1.95, -0.95) -- (closeup.north west);
		\draw[line width=0.25mm, blue] (1.95, -1.55) -- (closeup.south west);
		
		\begin{scope}[shift=(closeup.south west),x={(closeup.south east)},y={(closeup.north west)}]
        	\node[boxstyle,callout relative pointer={(0.08, 0.17)}] at (0.08,0.53) {Computer};
        	\node[boxstyle,callout relative pointer={(0.07, 0.)}] at (0.07,0.22) {LiDAR};
        	\node[boxstyle,text width=1.15cm,callout relative pointer={(-0.05, 0.35)}] at (.70,0.14) {RGB-D Cameras};
        	\node[boxstyle,text width=1.05cm,callout relative pointer={(0.02, 0.15)}] at (.85,0.44) {Thermal Camera};
    	\end{scope}
  \end{tikzpicture}}
  \caption{UAV system setup and hardware design.}
  \label{fig:hardware}
  \vspace{-1.5em}
\end{figure}
An overview of our UAV system, based on the commercially available DJI Matrice 210 v2 platform, is shown in \reffig{fig:hardware}.
We use an Intel Bean Canyon NUC8i7BEH with Core i7-8559U processor and \SI{32}{\giga\byte} of RAM as the onboard computer. A Google EdgeTPU connects to the NUC over USB 3.0 and accelerates CNN inference together with the Iris Plus Graphics 655 iGPU of the main processor.
An Ouster OS0-128 3D-LiDAR with 128 beams, \SI{360}{\degree} horizontal, and \SI{90}{\degree} vertical opening angles provides range measurements for 3D perception and odometry.
For visual perception, our UAV additionally carries two Intel RealSense D455 RGB-D cameras, mounted on top of each other to increase the vertical field-of-view, and a FLIR ADK thermal camera for, \eg person detection in search and rescue scenarios.

\subsection{Semantic Perception}
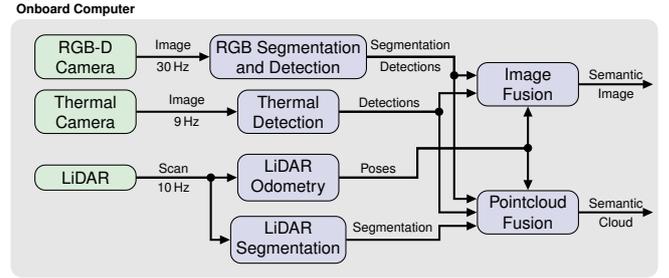
\begin{figure}[t]
  \centering
  \resizebox{1.0\linewidth}{!}{%
\begin{tikzpicture} 
[content_node/.append style={font=\sffamily,minimum size=1.5em,minimum width=6em,draw,align=center,rounded corners,scale=0.65},
label_node/.append style={font=\sffamily,scale=0.5},
group_node/.append style={font=\sffamily,dotted,align=center,rounded corners,inner sep=1em,thick},>={Stealth[inset=0pt,length=4pt,angle'=45]}]
\tikzset{junction/.append style={circle, fill=black, minimum size=3pt, draw}}

\definecolor{red}{rgb}     {0.5,0.0,0.0}
\definecolor{green}{rgb}   {0.0,0.5,0.0}
\definecolor{blue}{rgb}    {0.0,0.0,0.5}
\definecolor{grey}{rgb}    {0.5,0.5,0.5}

\draw[thick, rounded corners, grey!20!white,fill] (-4.0,1.5) -- (4.75,1.5) -- (4.75,5.0) -- (-4.0,5.0) -- cycle;

\node(Camera)[content_node,fill=green!15!white] at (-3.0,4.5) {RGB-D\\Camera};
\node(Thermal_Camera)[content_node,fill=green!15!white] at (-3.0,3.75) {Thermal\\Camera};
\node(LIDAR)[content_node,fill=green!15!white] at (-3.0,2.85) {LiDAR};

\node(RGB_Segm_Detection)[content_node,fill=blue!15!white] at (-.25,4.5) {RGB Segmentation\\and Detection};
\node(Thermal_Detection)[content_node,fill=blue!15!white] at (-.25,3.75) {Thermal\\Detection};
\node(LiDAR_Odom)[content_node,fill=blue!15!white] at (-.25,2.85) {LiDAR\\Odometry};
\node(LiDAR_Segm)[content_node,fill=blue!15!white] at (-.25,2.0) {LiDAR\\Segmentation};

\node(Image_Fusion)[content_node,fill=blue!15!white] at (3,4.125) {Image\\Fusion};
\node(Pointcloud_Fusion)[content_node,fill=blue!15!white] at (3,2.375) {Pointcloud\\Fusion};

\draw[->, thick] (Camera) -- node[label_node,midway,below] {\SI{30}{\hertz}} node[label_node,midway,above] {Image} (RGB_Segm_Detection);
\draw[->, thick] (Thermal_Camera) -- node[label_node,midway,below] {\SI{9}{\hertz}} node[label_node,midway,above] {Image} (Thermal_Detection);
\draw[->, thick] (LIDAR) -- (LiDAR_Odom);
\draw[->,thick] (LIDAR) -- node[label_node,midway,below] {\SI{10}{\hertz}} node[label_node,midway,above] {Scan} (LIDAR -| -1.3,-0.75) -- (-1.3,1.0 |- LiDAR_Segm.180) -- (LiDAR_Segm);

\draw[->, thick] (RGB_Segm_Detection) -- (RGB_Segm_Detection -| 2.,3.5) -- (2.,2.375 |- Image_Fusion.170) -- (Image_Fusion.170);
\draw[->, thick] (Thermal_Detection) -- (Thermal_Detection -| 1.8,3.5) -- (1.8,2.375 |- Image_Fusion.190) -- (Image_Fusion.190);

\draw[->, thick] (RGB_Segm_Detection) -- node[label_node,midway,above]{Segmentation} node[label_node,midway,below] {Detections} (RGB_Segm_Detection -| 2.,3.5) -- (2.,2.375 |- Pointcloud_Fusion.165) -- (Pointcloud_Fusion.165);
\draw[->, thick] (Thermal_Detection) -- node[label_node,midway,above] {Detections} (Thermal_Detection -| 1.8,3.5) -- (1.8,2.375 |- Pointcloud_Fusion.180) -- (Pointcloud_Fusion.180);
\draw[->, thick] (LiDAR_Segm) -- node[label_node,midway,above] {Segmentation} (LiDAR_Segm -| 1.8,3.5) -- (1.8,2.375 |- Pointcloud_Fusion.195) -- (Pointcloud_Fusion.195);
\draw[->, thick] (LiDAR_Odom) -- node[label_node,midway,above] {Poses} (LiDAR_Odom -| 1.5,2.375) -- ++(0.0, 0.4) -| (Pointcloud_Fusion.90);
\draw[->, thick] (LiDAR_Odom) -- (LiDAR_Odom -| 1.5,2.375) -- ++(0.0, 0.4) -| (Image_Fusion.270);

\draw[->, thick] (Image_Fusion) -- node[label_node,midway,below] {Image} node[label_node,midway,above] {Semantic} ++(1.7, 0.0);
\draw[->, thick] (Pointcloud_Fusion) -- node[label_node,midway,below] {Cloud} node[label_node,midway,above] {Semantic} ++(1.7, 0.0);

\node(junct_0)[junction, above right=0.0cm and 1.01cm of LIDAR.east, anchor=center, scale=0.3]{};
\node(junct_1)[junction, above left=0.0cm and .31cm of Image_Fusion.170, anchor=center, scale=0.3]{};
\node(junct_2)[junction, above right=0.0cm and 1.355cm of Thermal_Detection.east, anchor=center, scale=0.3]{};
\node(junct_3)[junction, below right=0.57cm and 0.0cm of Image_Fusion.270, anchor=center, scale=0.3]{};

\node(PC_Group_Label)[label_node,anchor=south west] at (-4.0,5.0) {\textbf{Onboard Computer}};
\end{tikzpicture}
}
  \caption{Perception system overview.}
  \label{fig:system}
  \vspace{-1.5em}
\end{figure}

An overview of the proposed architecture for multi-modal semantic perception is given in \reffig{fig:system}. We detail individual components in the following.
\subsubsection*{Image Segmentation}
We employ the DeepLabv~3+~\cite{deeplabv3plus2018} architecture with MobileNet~v3~\cite{mobilenetv32019} backbone optimized for Google EdgeTPU Accelerator~\cite{edgetpu_usb} for semantic segmentation. We train the model on the Mapillary Vistas Dataset~\cite{MVD2017}, reducing the labels to the 15 most relevant classes for the envisaged UAV tasks (\cf \reffig{fig:img_semantics}). We use an input image size of 849$\,\times\,$481~px during inference, fitting the 16:9 aspect ratio of our camera.
\subsubsection*{Object Detection}
The recent MobileDet architecture~\cite{xiong_mobiledets_2021} is the basis for our object detection. We train the RGB detector on the COCO dataset~\cite{lin_coco_2014} for \textit{person}, \textit{vehicle}, and \textit{bicycle} classes with an input resolution of 848$\,\times\,$480~px. The thermal object detector uses the same architecture taking one-channel 8-bit gray-scale thermal images at the full camera resolution of 640$\,\times\,$512~px as input.
We enable automatic gain correction (AGC) for the thermal camera for compatibility with the FLIR ADAS dataset~\cite{flir_dataset}. AGC adapts and scales the 16-bit raw images to 8-bit exploiting the full 8-bit value range. The network is trained on the ADAS dataset, with annotations for \textit{persons}, \textit{vehicles}, and \textit{bicycles}.
\subsubsection*{Point Cloud Segmentation}
We adopt the projection-based SalsaNext architecture~\cite{cortinhal_salsanext_2020}, pretrained on the large-scale SemanticKITTI~\cite{behley2019iccv} dataset that takes advantage of the image-like structure of LiDAR measurements. Subsampling of the OS0 scans by a factor of two in vertical and horizontal directions leads to a network input resolution of 64$\,\times\,$512. 
The input channels are range, $x$-, $y$-, $z$-coordinate, and intensity, normalized with the mean and standard deviation of the training dataset. Our LiDAR has a significantly larger vertical field-of-view of \SI{90}{\degree}, compared to the \SI{26.9}{\degree} vertical opening angle of the SemanticKITTI dataset. A different laser wavelength also changes the characteristics of intensity and reflections. 
Hence, we adjust the normalization parameters for $z$-coordinate and intensity to facilitate the domain adaptation between training and observed data according to the statistics of the test data captured with our sensor setup. The $x$- and $y$-coordinate normalization parameters remain the same, as the horizontal field-of-view is identical (\SI{360}{\degree}) for both sensors.
\reffig{fig:pointcloud_semantics} highlights improvements of the segmentation results through the adaptation of the normalization parameters.
\subsubsection*{Inference Accelerators}
We run the CNN model inference on two different accelerators onboard the UAV PC: The Google EdgeTPU~\cite{edgetpu_usb}, attached as an external USB device, and the integrated GPU (iGPU) included in most modern processors which is otherwise unused in our system.
The EdgeTPU supports network inference via TensorFlow-lite~\cite{abadi2016tensorflow} and requires quantization of the network weights and activations to 8-bit~\cite{quantization_2018}. The iGPU supports inference via the Intel OpenVINO framework~\cite{openvino_framework} in 16- or 32-bit floating-point precision.
\subsection{Multi-Modality Fusion}
\label{sec:mm_fusion}
We adopt a projection-based approach to fuse semantic class scores from image and point cloud. 
Projection onto the image plane requires the transformation of LiDAR points into the respective camera coordinate frame. As LiDAR and cameras operate with different frame rates, the motion between the respective capture times has to be taken into account. The full transformation chain $\vec{T}$ from LiDAR to camera frame is:
\begin{align}
\vec{T} &= {}^{\text{cam}}\vec{T}_{\text{base}} {}^{\text{base}_{t_c}}\vec{T}_{\text{base}_{t_l}} {}^{\text{base}}\vec{T}_{\text{LiDAR}}\,,
\end{align}
using the continuous-time trajectory of the UAV base frame estimated by the LiDAR odometry. Thus, the transformation chain models perspective changes between LiDAR and camera that occur due to dynamic UAV motions.

Bilinear interpolation at the projected point location gives the semantic class scores from image segmentation. A linear combination of the image score and the point score yields the fused class score:
\begin{align}
\vec{c}_\text{fused} &= (1-w_\text{img}) \vec{c}_\text{LiDAR} + w_\text{img} \vec{c}_\text{img}\,,
\end{align}
with $w_\text{img} \in [0,1]$ and $\vec{c}\in\mathbb{R}^{C}$ the vectors of class scores of the CNN output after soft-max of the $C=15$ classes used in this work (\cf \reffig{fig:img_semantics} (e)).

Furthermore, if a projected point falls inside a detection box in either thermal or color images, the detected class is included in the result. We base the detection weight $w_\text{det}$ on the detector score multiplied with a Gaussian factor with mean at the bounding box center and standard deviation of half the bounding box width resp. height. Again, a linear combination fuses both estimates:
\begin{align}
\vec{c}_\text{fused\_det} &= \left(1 - w_\text{det}\right) \vec{c}_\text{fused} + w_\text{det} \vec{c}_\text{det}\,.
\end{align}

Simple projection of all points into the rectangular bounding box will falsely label points in the background as the detected class (\cf \reffig{fig:pointcloud_semantics_person}). To alleviate this issue, points are clustered\wrt their distance in the camera frame before detection fusion. We include only points within the \SI{25}{\percent} quantile of distances per cluster to focus on foreground objects.
The final segmented point cloud includes the full class score vector and the argmax class color per point.

We proceed similarly with the initial image segmentation and detections from RGB and thermal cameras.
The RGB-D depth enables projection from RGB to thermal image and temporal smoothing provides a more coherent fused segmentation. For temporal fusion, we project the previous image at time $t-1$ with its depth into the current frame at time $t$ and perform exponential smoothing:
\begin{align}
\vec{c}_{\text{smoothed\_img}_{t}} &= \vec{\alpha} \circ \vec{c}_{\text{img}_{t}} + \left(\vec{1} - \vec{\alpha}\right) \circ \vec{c}_{\text{fused\_img}_{t-1}}\,,\\
\vec{c}_{\text{fused\_img}_t} &= \left(1 - w_\text{det}\right) \vec{c}_{\text{smoothed\_img}_{t}} + w_\text{det} \vec{c}_{\text{det}_t}\,,
\end{align}
with the coefficient-wise product $\circ$.

The smoothing weights $\vec{\alpha}$ differ between the individual semantic classes. For (potentially) dynamic foreground objects, such as \textit{persons} and \textit{vehicles}, less smoothing is applied (\ie larger coefficients in $\vec{\alpha}$) than for static structures such as buildings and roads. This reduces temporal jitter in the segmentation significantly\wrt the initial CNN output and still permits to follow the movement of dynamic objects.

\subsection{Semantic Mapping}
MARS LiDAR Odometry~\cite{quenzel2021mars} provides poses to integrate all augmented point clouds within a common map. A uniform grid subdivides the space into cubic volume elements (voxels). Since a dense voxel grid may require a prohibitively large amount of memory although only sparse access occurs, we use sparse voxel hashing. Each voxel fuses all points in its vicinity probabilistically. Additionally, we compute the mean position.
Our fusion scheme follows the reasoning of SemanticFusion~\cite{mccormac_semanticfusion_2017} to use Bayes' Rule assuming independence between semantic segmentations $P\left( l_i \lvert X_k\right)$ for the augmented point cloud $X_k$ with label $l_i$ for class $i$:
\begin{align}
P( l_i \lvert X_{1:k} ) &=  \frac{P\left(l_i \lvert X_{1:k-1}\right)  P \left( l_i \lvert X_k \right)}{\sum_i P\left( l_i \vert X_{1:k-1} \right) P\left( l_i \vert X_k\right)}.\label{eq:bayes}
\end{align}
A naive implementation, as in SemanticFusion, suffers from numerical instability due to finite precision of the multiplication result. In practice, this leads to all class probabilities being close to zero, \eg when $P\left( l_i \lvert X_k\right)\approx 1$, $P\left( l_j \lvert X_{k+1}\right)\approx 1$ and $P\left( l_i \lvert X_{k+1}\right)\approx 0$, $P\left( l_j \lvert X_k\right)\approx 0$ both class-wise products will be almost zero. This results in a loss of information even after the application of the normalization term and needs continuous reinitialization.
Hence, we implement \refeq{eq:bayes} with log-probabilities:
\begin{align}
L_{i,1:k-1} &= \log(P\left(l_i \lvert X_{1:k-1}\right)),\\
L_{i,k} &= \log(P \left( l_i \lvert X_k \right),\\
C_{i,1:k} &= L_{i,1:k-1}+L_{i,k},\\
M_{1:k} &= \max_i\left(C_{i,1:k}\right),\\
N_{1:k} &= \log\left(1+\sum_j{\exp^{C_{j,1:k}-M_{1:k}}}\right),\\
L_{i,1:k} &= C_{i,1:k} - \left(M_{1:k} + N_{1:k}\right),\\
P\left(l_i \lvert X_{1:k}\right) &= \exp^{L_{i,1:k}}.
\end{align}
Voxels now store $L_{i,1:k}$ instead of $P\left( l_i \lvert X_{1:k}\right)$. 
Ideally, we would directly fuse network outputs before soft-max to save additional $\exp$ and $\log$ evaluations, but since the individual outputs may be arbitrarily scaled, this step is necessary.

An infinite time horizon of the semantic map, fusing all scans, may not be necessary or wanted---depending on the use-case, e.g. for global vs. local planning. Hence, we employ a fixed-size double-ended queue (deque) per voxel for a shorter time horizon of $n$ scans that merges all points per scan. Fusion of per-scan log-probabilities yields the voxels' class probabilities. Older scans are either removed completely or fused into the infinite time horizon estimate.
 
\section{Evaluation}
\label{sec:Evaluation}
We first evaluate inference speed and computational efficiency of the employed CNN models and then show results from our outdoor UAV flights in an urban environment.

\subsection{CNN Model Efficiency}
In real-time systems with limited computational resources, such as UAVs, efficiency is of key importance and resources need to be distributed with care between the different system components. Semantic perception, while important for many high-level tasks, has less severe real-time constraints than, \eg flight control or odometry. It is thus important that the CNN inference uses as few CPU resources as possible to not interfere with the hard real-time constraints of low-level control, localization, and state estimation.
For this, we analyze the CPU load of the employed CNNs for object detection and segmentation, depending on the used accelerator. Although the main computational load of inference is distributed to a dedicated accelerator (EdgeTPU or iGPU), the preparation of input data, data transfer, and post-processing require CPU resources.
This is handled with differing degrees of efficiency\wrt CPU load and depends on the in- and output frame rate, as shown in \reffig{fig:inf_cpu_fps}. Models running on the EdgeTPU produce lower CPU load in all cases while achieving higher or equivalent maximum frame rates. \reftab{tab:model_runtime} shows the average inference latency per model. The LiDAR segmentation is only executed on the iGPU, as the \textit{pixel-shuffle} layer from SalsaNext~\cite{cortinhal_salsanext_2020} is not supported by the EdgeTPU and the model thus cannot be converted to the required 8-bit quantized format.

For the following experiments, we choose to run the image CNNs on the EdgeTPU, while the LiDAR segmentation runs on the iGPU. \reftab{tab:multi_inf_cpu_fps} shows the average computational load and output rate for different combinations of CNNs. As to be expected, the maximum achievable output frame rate drops and CPU load increases with a growing number of vision models used. 
The computation of RGB segmentation and detection as well as thermal detection achieves an average frame rate of \SI{12.6}{\hertz} at a CPU load of about \SI{60}{\percent}. The inclusion of the image fusion module almost doubles the CPU utilization while the frame rate drops to \SI{9.9}{\hertz}.
This is due to the transformations and projections necessary to calculate at image resolution for temporal smoothing and to include thermal detection into the fused image segmentation.
The total CPU usage for the fusion of both image and point cloud semantics sums up to about \num{2} CPU cores with an output rate of around \SI{9}{\hertz}. 

Reducing the input frequency to the semantic segmentation and detection can free additional resources for other system components if necessary while still providing semantic image and point cloud \eg at \SIrange[range-units=single]{1}{5}{\hertz}---sufficient for many high-level tasks like planning or keyframe-based mapping.
 
\begin{figure}
  \centering
  \resizebox{1.0\linewidth}{!}{%
\begin{tikzpicture}
\definecolor{red}{rgb}     {0.6350, 0.0780, 0.1840}
\definecolor{green}{rgb}   {0.4660, 0.6740, 0.1880}
\definecolor{blue}{rgb}    {0, 0.4470, 0.7410}
\definecolor{orange}{rgb}    {0.8500, 0.3250, 0.0980}
\begin{axis}[
    name=plot1,
    axis x line=bottom,
    axis y line=left,
    xticklabel={$\mathsf{\pgfmathprintnumber{\tick}}$},
    yticklabel={$\mathsf{\pgfmathprintnumber{\tick}}$},
    extra y ticks={20,40,...,120},
    extra tick style={grid=major},
    x label style={font=\small\sffamily, at={(axis description cs:0.5,-0.08)},anchor=north},
    y label style={font=\small\sffamily, at={(axis description cs:-0.08,.5)},anchor=south},
    xlabel={FPS (Hz)},
    ylabel={CPU load (\%)},
    ymin=0,
    xmin=0,
    xmax=31,
    line width=1.5pt,
    ]
    
 \addplot+[color=red, line width=2pt, solid, mark=*, mark options={solid, scale=1.25}]
 plot [error bars/.cd, y dir = both, y explicit]
 table[x=FPS, y=CPU, y error plus=STD, y error minus=STDMIN]{data/edgetpu_segm.txt};
 \label{plot:rgbseg_tpu}

 \addplot+[color=red, line width=2pt, dashed, mark=o, mark options={solid, scale=1.35, line width=1.5pt}]
 plot [error bars/.cd, y dir = both, y explicit]
 table[x=FPS, y=CPU, y error plus=STD, y error minus=STDMIN]{data/igpu_segm.txt};
 \label{plot:rgbseg_igpu}
 \end{axis}
 
 \begin{axis}[
    name=plot2,
    at=(plot1.below south west), anchor=above north west,
    axis x line=bottom,
    axis y line=left,
    xticklabel={$\mathsf{\pgfmathprintnumber{\tick}}$},
    yticklabel={$\mathsf{\pgfmathprintnumber{\tick}}$},
    extra y ticks={20,40,...,120},
    extra tick style={grid=major},
    x label style={font=\small\sffamily, at={(axis description cs:0.5,-0.08)},anchor=north},
    y label style={font=\small\sffamily, at={(axis description cs:-0.08,.5)},anchor=south},
    xlabel={FPS (Hz)},
    ylabel={CPU load (\%)},
    ymin=0,
    xmin=0,
    xmax=31,
    line width=1.5pt,
    ]
 
 \addplot+[color=green, line width=2pt, solid, mark=x, mark options={solid, scale=1.75, line width=1.5pt}]
 plot [error bars/.cd, y dir = both, y explicit]
 table[x=FPS, y=CPU, y error plus=STD, y error minus=STDMIN]{data/edgetpu_rgb_det.txt};
 \label{plot:rgbdet_tpu}

 \addplot+[color=green, line width=2pt, dashed, mark=x, mark options={solid, scale=1.75, line width=1.5pt}]
 plot [error bars/.cd, y dir = both, y explicit]
 table[x=FPS, y=CPU, y error plus=STD, y error minus=STDMIN]{data/igpu_rgb_det.txt};
 \label{plot:rgbdet_igpu}

 \end{axis}
 
 \begin{axis}[
    name=plot3,
    at=(plot2.right of south east), anchor=left of south west,
    axis x line=bottom,
    axis y line=left,
    xticklabel={$\mathsf{\pgfmathprintnumber{\tick}}$},
    yticklabel={$\mathsf{\pgfmathprintnumber{\tick}}$},
    extra y ticks={10,20,...,60},
    extra tick style={grid=major},
    x label style={font=\small\sffamily, at={(axis description cs:0.5,-0.08)},anchor=north},
    y label style={font=\small\sffamily, at={(axis description cs:-0.08,.5)},anchor=south},
    xlabel={FPS (Hz)},
    ylabel={CPU load (\%)},
    ymin=0,
    xmin=0,
    xmax=11,
    line width=1.5pt,
    ]
 
 \addplot+[color=blue, line width=2pt, solid, mark=diamond*, mark options={solid, scale=1.25}]
 plot [error bars/.cd, y dir = both, y explicit]
 table[x=FPS, y=CPU, y error plus=STD, y error minus=STDMIN]{data/edgetpu_thermal_det.txt};
 \label{plot:thermaldet_tpu}

 \addplot+[color=blue, line width=2pt, dashed, mark=diamond, mark options={solid, scale=1.5, line width=1.5pt}]
 plot [error bars/.cd, y dir = both, y explicit]
 table[x=FPS, y=CPU, y error plus=STD, y error minus=STDMIN]{data/igpu_thermal_det.txt};
 \label{plot:thermaldet_igpu}

 \addplot+[color=orange, line width=2pt, dashed, mark=triangle, mark options={solid, scale=1.5, line width=1.5pt}]
 plot [error bars/.cd, y dir = both, y explicit]
 table[x=FPS, y=CPU, y error plus=STD, y error minus=STDMIN]{data/igpu_lidar_segm.txt};
 \label{plot:lidarsegm_igpu}
 \end{axis}

\matrix[
    matrix of nodes,
    ampersand replacement=\&,
    anchor=north west,
    below right = 0.7cm and 0.6cm of plot1.north east,
    draw,%
    inner sep=0.3em,
    column 1/.style={nodes={anchor=center}},
    column 2/.style={nodes={anchor=west},font=\Large},
    draw
  ]
  {
    \ref{plot:rgbseg_tpu}\& RGB segmentation (EdgeTPU)\\
    \ref{plot:rgbseg_igpu}\& RGB segmentation (iGPU)\\
    \ref{plot:rgbdet_tpu}\& RGB detector (EdgeTPU)\\
    \ref{plot:rgbdet_igpu}\& RGB detector (iGPU)\\
    \ref{plot:thermaldet_tpu}\& Thermal detector (EdgeTPU)\\
    \ref{plot:thermaldet_igpu}\& Thermal detector (iGPU)\\
    \ref{plot:lidarsegm_igpu}\& LiDAR segmentation (iGPU)\\
    };
    
\end{tikzpicture}
}
  \caption{CPU load of the CNN inference of different models depending on the used accelerator and the output frame rate. The iGPU (dashed lines) results in higher CPU load than the EdgeTPU (solid lines) for all models.}
  \label{fig:inf_cpu_fps}
  \vspace{-.5em}
\end{figure}
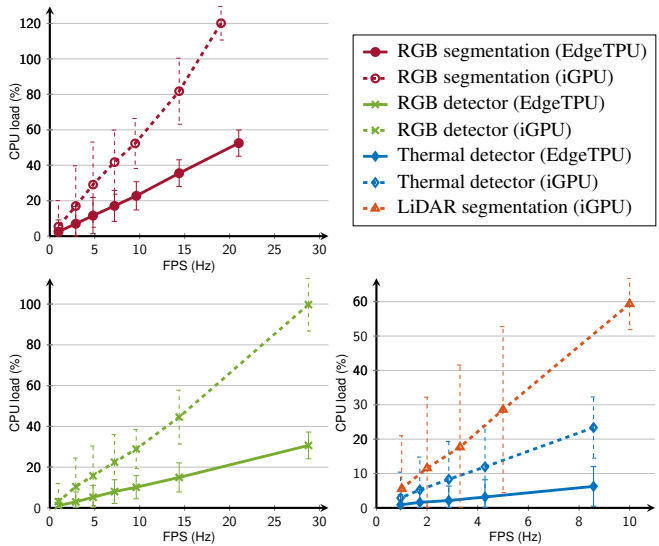
\begin{table}[t]
\caption{Average inference time on the resp. accelerator.} 
\label{tab:model_runtime}
\centering
\begin{threeparttable}
\begin{tabular}{l|c|c}
  \toprule %
  Model & EdgeTPU & iGPU\\
  \midrule %
  RGB Segmentation & \SI{40.5}{\milli\second} & \SI{50.0}{\milli\second}\\
  RGB Detection & \SI{17.5}{\milli\second} & \SI{24.0}{\milli\second} \\
  Thermal Detections & \SI{12.0}{\milli\second} & \SI{18.0}{\milli\second} \\
  LiDAR Segmentation & - & \SI{32.0}{\milli\second} \\
  \bottomrule %
\end{tabular}
\end{threeparttable}
\vspace{-1.2em}
\end{table}
\begin{table}[t]
\caption{Average CPU load and output frame rate of different model combinations. Image segmentation and detection models run on EdgeTPU and point cloud segmentation on iGPU.} 
\label{tab:multi_inf_cpu_fps}
\centering
\vspace{-1em}
\hspace{-3.5em}
\setlength{\extrarowheight}{.1em}
\setlength{\tabcolsep}{0.7em} 
\begin{threeparttable}
\begin{tabular}{ccccccccc}
  \rot{point cloud fusion} & \rot{image fusion} & \rot{point cloud segm.} & \rot{thermal dets.} & \rot{RGB dets.} & \rot{RGB segm.} & \rot{CPU load} & \rot{avg. FPS (image)} & \rot{avg. FPS (cloud)}\\
  \midrule %
  - & - & - & - & - & \checkmark & \multicolumn{1}{|c|}{\SI{52.5}{\percent}} & \multicolumn{1}{c|}{\SI{21.0}{\hertz}} & - \\
  - & - & - & - & \checkmark & \checkmark & \multicolumn{1}{|c|}{\SI{54.2}{\percent}} & \multicolumn{1}{c|}{\SI{13.2}{\hertz}} & - \\
  - & - & - & \checkmark & \checkmark & \checkmark & \multicolumn{1}{|c|}{\SI{57.3}{\percent}} & \multicolumn{1}{c|}{\SI{12.6}{\hertz}} & - \\
  - & \checkmark & - & \checkmark & \checkmark & \checkmark & \multicolumn{1}{|c|}{\SI{120.6}{\percent}} & \multicolumn{1}{c|}{\SI{9.9}{\hertz}} & - \\
  - & - & \checkmark & \checkmark & \checkmark & \checkmark & \multicolumn{1}{|c|}{\SI{116.6}{\percent}} & \multicolumn{1}{c|}{\SI{12.6}{\hertz}} & \SI{10.0}{\hertz} \\
  - & \checkmark & \checkmark & \checkmark & \checkmark & \checkmark & \multicolumn{1}{|c|}{\SI{180.0}{\percent}} & \multicolumn{1}{c|}{\SI{9.5}{\hertz}} & \SI{10.0}{\hertz} \\
  \checkmark & \checkmark & \checkmark & \checkmark & \checkmark & \checkmark & \multicolumn{1}{|c|}{\SI{204.3}{\percent}} & \multicolumn{1}{c|}{\SI{8.9}{\hertz}} & \SI{9.5}{\hertz} \\
  \bottomrule %
\end{tabular}
\end{threeparttable}
\vspace{-1.4em}
\end{table}

\subsection{Outdoor Experiments}
\begin{figure}[t]
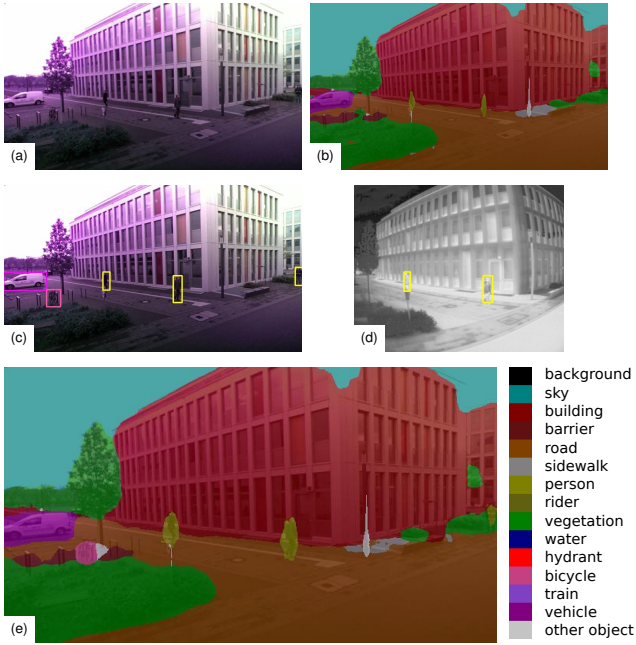

	\centering
		\begin{tikzpicture}
			\node[inner sep=0,anchor=north west] (image1) at (0,0) {\includegraphics[height=2.2cm]{figures/scenes/scene1_input_image.jpg}};
			\node[inner sep=0,anchor=north west,xshift=0.1cm] (image2) at (image1.north east) {\includegraphics[height=2.2cm]{figures/scenes/scene1_rgb_segm.jpg}};
			\node[inner sep=0,anchor=north west,yshift=-0.2cm] (image3) at (image1.south west) {\includegraphics[height=2.2cm]{figures/scenes/scene1_rgb_dets.jpg}};
			\node[inner sep=0,anchor=north,yshift=-0.2cm] (image4) at (image2.south) {\includegraphics[height=2.2cm]{figures/scenes/scene1_thermal_dets.jpg}};
			\node[inner sep=0,anchor=north west,yshift=-0.2cm] (image5) at (image3.south west) {\includegraphics[width=0.75\linewidth]{figures/scenes/scene1_fused_segm.jpg}};
			\node[inner sep=0,anchor=north west,xshift=0.15cm] (image6) at (image5.north east) {\includegraphics[width=0.19\linewidth]{figures/scenes/legend_mapillary_obj.pdf}};
			
			\node[label,scale=0.75, anchor=south west, rectangle, fill=white, align=center, font=\scriptsize\sffamily] (n_0) at (image1.south west) {(a)};
			\node[label,scale=0.75, anchor=south west, rectangle, fill=white, align=center, font=\scriptsize\sffamily] (n_1) at (image2.south west) {(b)};
			\node[label,scale=0.75, anchor=south west, rectangle, fill=white, align=center, font=\scriptsize\sffamily] (n_2) at (image3.south west) {(c)};
			\node[label,scale=0.75, anchor=south west, rectangle, fill=white, align=center, font=\scriptsize\sffamily] (n_3) at (image4.south west) {(d)};
			\node[label,scale=0.75, anchor=south west, rectangle, fill=white, align=center, font=\scriptsize\sffamily] (n_4) at (image5.south west) {(e)};
		\end{tikzpicture}
	\caption{Semantic interpretation of RGB and thermal images: (a) RGB input image, (b) segmentation and (c) detections. (d) thermal detections. (e) fused segmentation mask. Persons and bicycle, not or only partially segmented in the initial segmentation mask, are fully visible in the fused output.}
	\label{fig:img_semantics}
	\vspace{-1.3em}
\end{figure}

In \reffig{fig:img_semantics}, we show results of semantic image fusion for an exemplary scene from our test flights. \reffig{fig:img_semantics} (b) - (d) show the outputs of the individual CNNs. While the large structures are well segmented, the persons are only partially recognized. A bicycle and the person at the right image border are even missed altogether. The RGB detector recognizes all persons and the bicycle. The thermal detector confirms both person detections inside the thermal camera's field-of-view. The fused output segmentation mask (e) includes all detections together with the initial segmentation mask. All persons and the bicycle are clearly visible.

\begin{figure*}[t]
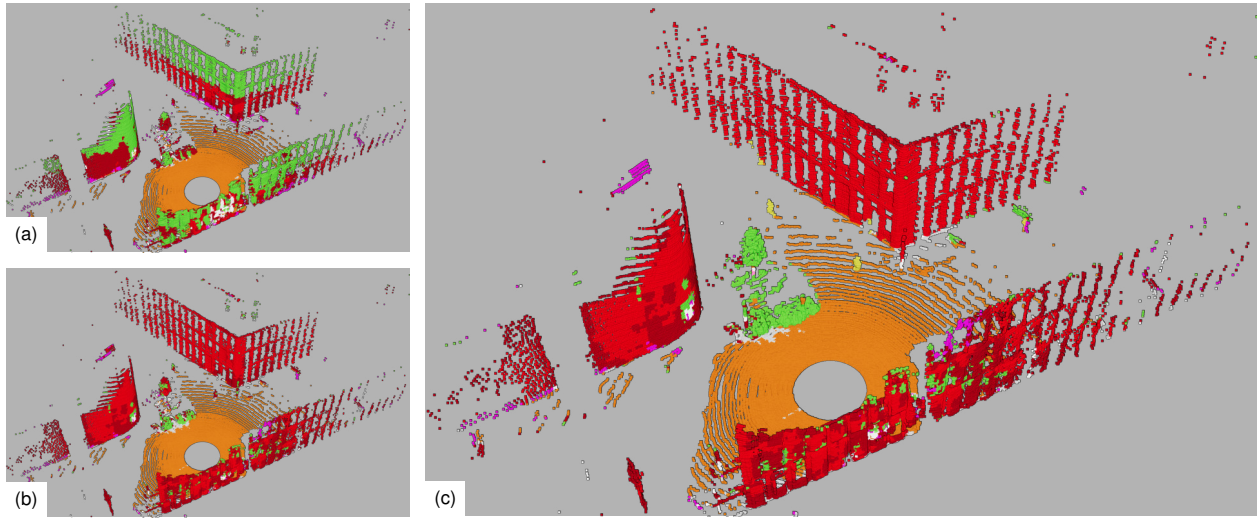

	\centering
		\begin{tikzpicture}
			\node[inner sep=0,anchor=north west,fill={white!70!black}] (image1) at (0,0) {\includegraphics[height=3.3cm,trim= 650 220 580 350,clip]{figures/scenes/scene1_lidar_raw_seg_e6.png}};
			\node[inner sep=0,anchor=north west,fill={white!70!black},yshift=-0.2cm] (image2) at (image1.south west) {\includegraphics[height=3.3cm,trim= 650 220 580 350,clip]{figures/scenes/scene1_adapted_lidarseg_e6.png}};
			\node[inner sep=0,anchor=north west,fill={white!70!black}, xshift=0.2cm] (image3) at (image1.north east) {\includegraphics[height=6.8cm,trim= 650 220 580 350,clip]{figures/scenes/scene1_adapted_fused_e6.png}};
			\node[label,scale=1, anchor=south west, rectangle, fill=white, align=center, font=\scriptsize\sffamily] (n_0) at (image1.south west) {(a)};
			\node[label,scale=1, anchor=south west, rectangle, fill=white, align=center, font=\scriptsize\sffamily] (n_1) at (image2.south west) {(b)};
			\node[label,scale=1, anchor=south west, rectangle, fill=white, align=center, font=\scriptsize\sffamily] (n_2) at (image3.south west) {(c)};
		\end{tikzpicture}
	\caption{LiDAR point cloud segmentation: (a) Initial segmentation without adaptation of z and intensity normalization parameters and (b) after adaptation to our dataset mean and std. (c) fused point cloud segmentation.
	After normalization adaptation, the CNN segments the large structures well within the LiDAR scan but misses small objects.
	Persons and small structures are well segmented in the fused output scan and buildings, vegetation, and the car exhibit less noise.}
	\label{fig:pointcloud_semantics}
	\vspace{-1em}
\end{figure*}

\reffig{fig:pointcloud_semantics} shows the point cloud segmentation results for the same scene. Without the adaptation of the normalization parameters, the top half of the buildings are misclassified as vegetation. This is likely due to differing vertical field-of-views of our and the trained LiDAR. In the KITTI dataset~\cite{geiger2012cvpr}, the FoV is only $\approx\,$\SI{3}{\degree} upwards and \SI{25}{\degree} downwards (compared to \SI{\pm45}{\degree} of our sensor). Furthermore, in SemanticKITTI the top of building structures is rarely visible while the LiDAR often measures treetops at the top end. 
After normalization adaptation, SalsaNext segments the large structures well within the LiDAR scan (\reffig{fig:pointcloud_semantics} (b)). \reffig{fig:pointcloud_semantics} (c) shows the fused point cloud segmentation, combining image segmentation and detections with the initial point cloud segmentation. Persons and small objects are well segmented in the output scan while buildings, vegetation, and the car exhibit less noise in the segmentation when inside the camera FoV. Independent of the normalization, the point cloud network does not detect persons, often misclassifying them as either vegetation or buildings. 

\begin{figure}
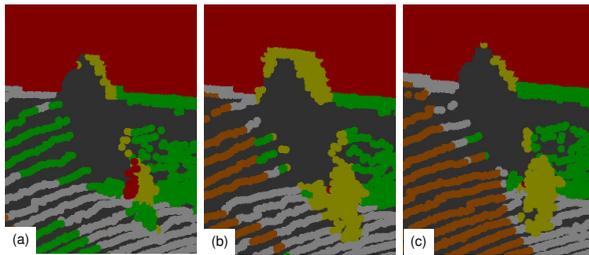

	\centering
		\begin{tikzpicture}
			\node[inner sep=0,anchor=north west] (image1) at (0,0) {\includegraphics[trim=53mm 16mm 158mm 95mm,clip,width=0.29\linewidth]{figures/person_imgsegm_only.png}};
			\node[inner sep=0,anchor=north west,xshift=0.1cm] (image2) at (image1.north east) {\includegraphics[trim=51mm 17mm 167mm 93mm,clip,width=0.29\linewidth]{figures/person_det.png}};
			\node[inner sep=0,anchor=north west,xshift=0.1cm] (image3) at (image2.north east) {\includegraphics[trim=50mm 20mm 155mm 90mm,clip,width=0.29\linewidth]{figures/person_det_cluster.png}};
			\node[label,scale=0.75, anchor=south west, rectangle, fill=white, align=center, font=\scriptsize\sffamily] (n_0) at (image1.south west) {(a)};
			\node[label,scale=0.75, anchor=south west, rectangle, fill=white, align=center, font=\scriptsize\sffamily] (n_1) at (image2.south west) {(b)};
			\node[label,scale=0.75, anchor=south west, rectangle, fill=white, align=center, font=\scriptsize\sffamily] (n_2) at (image3.south west) {(c)};
		\end{tikzpicture}
	\caption{Person segmentation included into point cloud using (a) image segmentation only, (b) additionally detection bounding boxes, and (c) clustering foreground points within the detections bounding boxes. After clustering, the person is completely segmented without adding additional misclassified points in the background.}
	\label{fig:pointcloud_semantics_person}
	\vspace{-1.5em}
\end{figure}

In \reffig{fig:pointcloud_semantics_person}, we show the process of fusing person detections from the RGB or thermal image into the point cloud. The initial segmentation, using only the points projected onto the image segmentation mask, is incomplete and slightly misaligned. The addition of points projected into the bounding box creates many false positives in the background. Our inclusion of only foreground points, after clustering\wrt the distance to the sensor (\cf \refsec{sec:mm_fusion}), results in the entire person being correctly labeled without adding additional mislabeled points in the background.

\begin{figure*}[t]
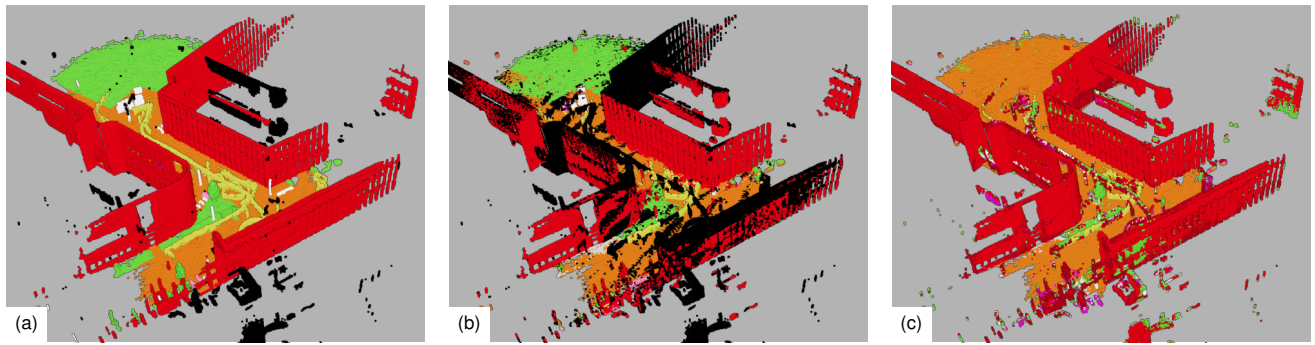

	\centering
		\begin{tikzpicture}
			\node[inner sep=0,anchor=north west,fill={white!70!black}] (image0) at (0,0) {\includegraphics[width=0.31\linewidth,trim= 500 0 1120 320,clip]{figures/scenes/ecmr_campus_gt.png}};
			\node[inner sep=0,anchor=north west,xshift=0.3cm,fill={white!70!black}] (image1) at (image0.north east) {\includegraphics[width=0.31\linewidth,trim= 500 0 1120 320,clip]{figures/scenes/ecmr_campus_camonly.png}};
			\node[inner sep=0,anchor=north west,xshift=0.3cm,fill={white!70!black}] (image2) at (image1.north east)  {\includegraphics[width=0.31\linewidth,trim= 500 0 1120 320,clip]{figures/scenes/ecmr_campus_fused.png}};
			\node[label,scale=1, anchor=south west, rectangle, fill=white, align=center, font=\scriptsize\sffamily] (n_0) at (image0.south west) {(a)};
			\node[label,scale=1, anchor=south west, rectangle, fill=white, align=center, font=\scriptsize\sffamily] (n_0) at (image1.south west) {(b)};
			\node[label,scale=1, anchor=south west, rectangle, fill=white, align=center, font=\scriptsize\sffamily] (n_1) at (image2.south west) {(c)};
		\end{tikzpicture}
	\caption{Semantic map of the longest outdoor flight. (a) manually annotated ground-truth. (b) map created from scans labeled by projected image segmentation only. (c) map created from fused semantic clouds. The camera-only map misses annotations due to the camera's limited FoV but depicts most classes more accurately, since the noisier raw point cloud segmentation is not included.}
	\label{fig:semantic_map}
	\vspace{-0.6em}
\end{figure*}

\begin{figure*}[t]
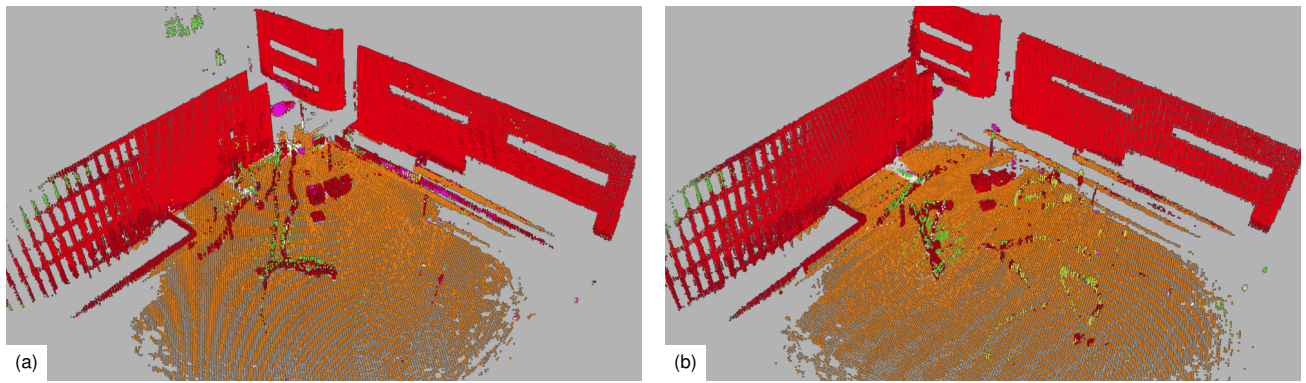

	\centering
		\begin{tikzpicture}
			\node[inner sep=0,anchor=north west,fill={white!70!black}] (image1) at (0, 0) {\includegraphics[width=0.47\linewidth,trim= 380 320 480 100,clip]{figures/scenes/map_second_flight_fused_e6.png}};
			\node[inner sep=0,anchor=north west,xshift=0.3cm,fill={white!70!black}] (image2) at (image1.north east)  {\includegraphics[width=0.47\linewidth,trim= 380 320 480 100,clip]{figures/scenes/map_third_flight_fused_e6.png}};
			\node[label,scale=1, anchor=south west, rectangle, fill=white, align=center, font=\scriptsize\sffamily] (n_0) at (image1.south west) {(a)};
			\node[label,scale=1, anchor=south west, rectangle, fill=white, align=center, font=\scriptsize\sffamily] (n_1) at (image2.south west) {(b)};
		\end{tikzpicture}
	\caption{Aggregated maps from fused semantic clouds of two further experiments with (a) the camera pointed away from and (b) towards people. Within the camera frustum, person detection works sufficiently well, while they are misclassified as barrier or vegetation elsewhere.}
	\label{fig:more_semantic_maps}
	\vspace{-1.7em}
\end{figure*}

\reffig{fig:semantic_map} depicts the resulting semantic map of the longest test flight with manually annotated semantic labels (a) and with scans either labeled from image segmentation (b) or from fused semantic point clouds (c). The camera-only map (b) misses annotations due to the camera's limited FoV but depicts most classes, such as persons, cars, or vegetation, more accurately since the noisier raw point cloud segmentation is not included. The tracks of moving persons are clearly visible in yellow in both maps. Only the track of the operator, who always stayed behind the UAV and thus was not visible in the camera, is not segmented (b) or mislabeled (c). The direct segmentation of persons or small structures in the LiDAR scans is very noisy due to domain adaptation issues with the employed CNN.
The aggregated maps from fused semantic clouds of two further experiments, shown in \reffig{fig:more_semantic_maps}, underline this issue. In the first test (a) the camera pointed away from people and in the second test (b) more towards the persons within the scene (pointing right\wrt the image). Within the camera frustum, person detection works sufficiently well, while they are misclassified as barrier or vegetation elsewhere.

To quantitatively evaluate the coherence of different point cloud segmentations, we calculate the intersection-over-union (IoU):
\begin{align}
IoU_c &= \frac{TP_c}{TP_c+FP_c+FN_c}
\end{align}
where TP, FP, and FN are the true positives, false positives, and false negatives, respectively.
We compare for each segmented point its $\argmax$ class against the corresponding aggregated voxel label and average per class over the whole dataset. \reftab{tab:iou} shows the results for all classes that occur for a significant number of points in our recorded data. We use the manually annotated aggregated semantic map with a voxel size of \SI{25}{\centi\metre} as ground-truth (cf. \reffig{fig:semantic_map}~(a)).
Applying the proposed adaptation of the normalization parameters significantly improves the segmentation of the building class, as the top half of the buildings are correctly labeled (\cf \reffig{fig:pointcloud_semantics}).
The fused semantic cloud improves the segmentation coherence for all classes, especially for persons and vegetation. Persons and small objects, such as bicycles, however, are only correctly labeled within the camera FoV.
The track of the operator, who always stayed behind the UAV and was not visible in the camera, is misclassified, significantly impacting the mIOU of the person class.
Results for the semantic cloud from the projected image segmentation and the fused semantic cloud evaluated for the reduced FoV of the camera show significantly improved mIOU values also for persons, cars, and other foreground objects.

The proposed semantic fusion thus successfully creates a coherently labeled 3D semantic interpretation for the global structure in the full \SI{360}{\degree} LiDAR FoV and for both global structure and small dynamic objects inside the camera FoV. To improve the accuracy for the difficult semantic classes in the entire FoV, label propagation could further be used for retraining the LiDAR segmentation.

\begin{table}[t]
\caption{Average IoU per class (in \si{\percent}) and mean IoU for different point cloud segmentation approaches measured against the manually annotated semantic map.} 
\label{tab:iou}
\centering
\renewcommand{\arraystretch}{1.3} %
\setlength{\tabcolsep}{0.17em} 
\begin{threeparttable}
\begin{tabular}{L{3.2cm}|cccccccc}
  \toprule
  Segmentation Method & Build. & Road & Veg. & Pers. & Bike & Car & Obj. & Mean\\
  \midrule %
  point cloud segmentation w/o adapted normalization & 40.4 & 73.3 & 3.1 & 0.2 & 0.0 & 2.7 & 2.2 & 17.4\\
  point cloud segmentation w. adapted normalization & 81.5 & 73.8 & 5.9 & 0.1 & 0.0 & 3.6 & 1.9 & 23.8\\
  fused semantic cloud & \textbf{82.9} & \textbf{76.0} & \textbf{20.2} & \textbf{5.8} & \textbf{1.6} & \textbf{6.4} & \textbf{4.9} & \textbf{28.3}\\
  \midrule
  fused semantic cloud (\textit{reduced to camera FoV}) & \textit{93.4} & \textit{66.2} & \textit{59.9} & \textit{34.0} & \textit{16.6} & \textit{37.6} & \textit{34.9} & \textit{48.9}\\
  semantic cloud from image overlay (\textit{camera FoV}) & \textit{92.5} & \textit{76.5} & \textit{76.7} & \textit{33.5} & \textit{17.1} & \textit{43.5} & \textit{39.7} & \textit{54.2}\\
  \bottomrule %
\end{tabular}
\end{threeparttable}
\vspace{-1.2em}
\end{table}

\section{Conclusion}
\label{sec:Conclusion}
In this work, we presented a UAV system for semantic image and point cloud analysis as well as multi-modal semantic fusion.
The inference of the lightweight CNN models runs onboard the UAV computer, employing an inference accelerator and the integrated GPU of the main processor for computation. 
The EdgeTPU performs inference in 8-bit quantized mode and showed more efficient CPU usage. The iGPU is more flexible, \eg to directly run pre-trained models, as it uses 16- or 32-bit floating-point precision and does not require model quantization.
We evaluated the system in real-world experiments in an urban environment. The semantic scene analysis provides a 2D image segmentation overlay and a 3D semantically labeled point cloud which is further aggregated into an allocentric semantic map.

The current point cloud segmentation suffers from domain adaptation issues since available large-scale training datasets stem from autonomous driving scenarios with different viewpoints and sensors more focused towards the ground. Thus, future work includes fine-tuning the laser segmentation with simulated data of the employed LiDAR scanner or on data recorded with the UAV and using label propagation to retrain networks with the aggregated semantic map. 

\bibliographystyle{IEEEtran}
\bibliography{literature}

\end{document}